\documentclass[runningheads]{llncs}

\usepackage{eccv}

\usepackage{eccvabbrv}

\usepackage{graphicx}
\usepackage{booktabs}
\usepackage{bbding}
\usepackage{amssymb}
\usepackage{pdfpages}
\usepackage{amsmath}
\usepackage{makecell}
\usepackage{arydshln}
\usepackage{caption}
\usepackage{threeparttable}
\usepackage{float}

\usepackage[accsupp]{axessibility}  

\usepackage[pagebackref,breaklinks,colorlinks,citecolor=eccvblue]{hyperref}

\usepackage{orcidlink}

\begin{document}

\title{MoD-SLAM: Monocular Dense Mapping for Unbounded 3D Scene Reconstruction} 

\titlerunning{MoD-SLAM}

\author{Heng Zhou\inst{1} \and
Zhetao Guo\inst{4} \and
Shuhong Liu\inst{2}\and
Lechen Zhang\inst{1}\and
Qihao Wang\inst{1}\and
Yuxiang Ren\inst{5}\and
Mingrui Li\inst{3*}}

\authorrunning{H. Zhou et al.}

\institute{Columbia University 
\email{\{hz2944,lechen.zhang,qw2414\}@columbia.edu}
\and The University of Tokyo 
\email{\{s-liu\}@isi.imi.i.u-tokyo.ac.jp}
\and Dalian University of Technology 
\email{\{2905450254\}@mail.dlut.edu.cn}
\and Cloudspace Technology
\email{\{gzt\}@yhsk.com.cn}
\and Dianjing Ciyuan Culture
\email{\{yuxiang.ren\}@dianjcy.com}}

\maketitle

\begin{abstract}
Monocular SLAM has received a lot of attention due to its simple RGB inputs and the lifting of complex sensor constraints. However, existing monocular SLAM systems are designed for bounded scenes, restricting the applicability of SLAM systems. To address this limitation, we propose MoD-SLAM, the first monocular NeRF-based dense mapping method that allows 3D reconstruction in real-time in unbounded scenes. Specifically, we introduce a Gaussian-based unbounded scene representation approach to solve the challenge of mapping scenes without boundaries. This strategy is essential to extend the SLAM application. Moreover, a depth estimation module in the front-end is designed to extract accurate priori depth values to supervise mapping and tracking processes. By introducing a robust depth loss term into the tracking process, our SLAM system achieves more precise pose estimation in large-scale scenes. Our experiments on two standard datasets show that MoD-SLAM achieves competitive performance, improving the accuracy of the 3D reconstruction and localization by up to 30\% and 15\% respectively compared with existing state-of-the-art monocular SLAM systems. 
  \keywords{3D Reconstruction \and Neural Radiance Field \and SLAM \and Monocular Depth Estimation \and Unbounded Scene}
\end{abstract}

\section{Introduction}

Dense visual Simultaneous Localization and Mapping (SLAM) is crucial in 3D computer vision, finding utility across various fields including autonomous driving, unmanned aerial vehicle, AR, etc. To make SLAM valuable in real-world applications, we want SLAM to be real-time, fast, accurate, and adaptable to scenes of different scales. Although traditional SLAM methods \cite{Campos2020ORBSLAM3AA,MurArtal2016ORBSLAM2AO,Zubizarreta2019DirectSM} can build maps in real-time, they can only generate sparse maps. Some works \cite{Sucar2020NodeSLAMNO,Zhi2019SceneCodeMD,Zubizarreta2019DirectSM} can generate dense maps to a certain extent, but both of them are generally too slow to build maps and can not be applied to large unbounded scenes. These constraints limit the application of SLAM.

In recent years, due to the development and advancement of Neural Radiance Field (NeRF) \cite{Mildenhall2020NeRFRS,Deng2021DepthsupervisedNF} and view synthesis \cite{Niemeyer2019DifferentiableVR,Sitzmann2020ImplicitNR,Barron2021MipNeRFAM} in 3D scene reconstruction, dense visual SLAM mapping has become feasible. Early NeRF-based SLAM systems such as iMAP \cite{Sucar2021iMAPIM} and NICE-SLAM \cite{Zhu2021NICESLAMNI} have achieved ideal results in dense mapping and tracking. However, they all have the problems of decreasing accuracy and scale drift in large scenes. Subsequent researches \cite{Zhang2023HISLAMMR,Zhang2023GOSLAMGO,Wang2023CoSLAMJC,Chung2022OrbeezSLAMAR,Rosinol2022NeRFSLAMRD} apply the signed distance function (SDF) \cite{Park2019DeepSDFLC} to reconstruct the 3D scene more accurately. Nevertheless, these approaches still present limited capability in unbounded scenes. 

To represent unbounded scenes, a commonly adopted strategy is to employ spatial warping methods that map unbounded space to bounded space. However, warping implies information compression, which presents a significant challenge to the accuracy of features, especially for monocular SLAM systems. NEWTON \cite{Matsuki2023NEWTONNV} is the first RGB-D NeRF-based SLAM system capable of mapping unbounded scenes. Nevertheless, it lacks support for RGB input and relies on Nerfstudio \cite{Tancik2023NerfstudioAM} in back-end processing to achieve reconstruction of unbounded scenes. Thus, the applicability of this system is significantly limited. 

\begin{figure}[t]
\centering
\includegraphics[width=120mm]{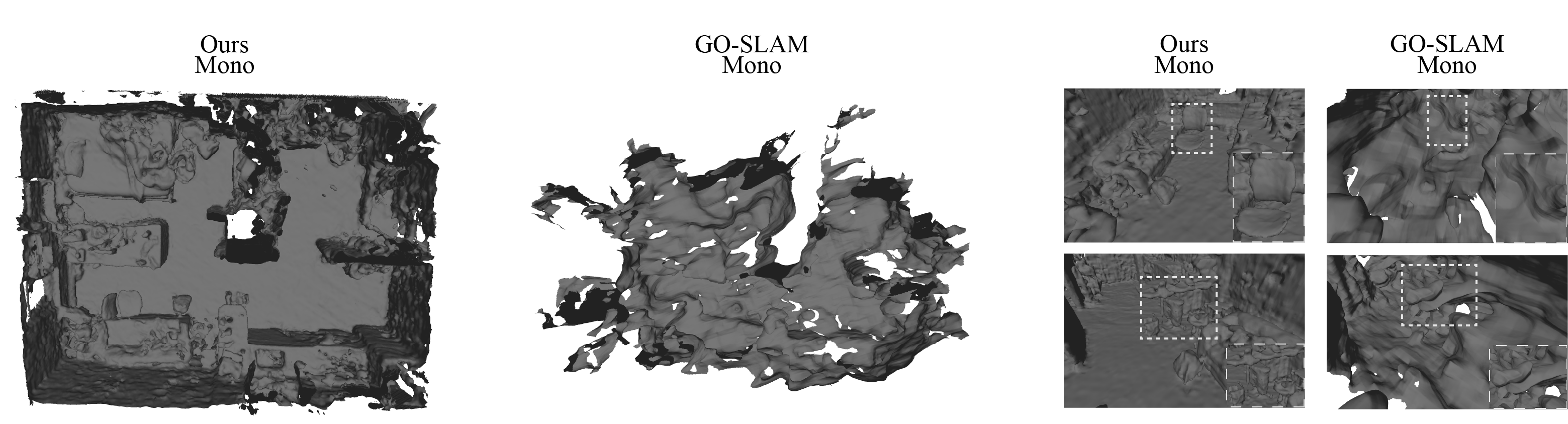}
\captionsetup{justification=justified,singlelinecheck=false}
\caption{\textbf{Reconstruction results on ScanNet 0207 \cite{Dai2017ScanNetR3}.} We present MoD-SLAM, a neural-based monocular dense mapping method. Our method shows a more powerful capability than existing state-of-the-art work: GO-SLAM \cite{Zhang2023GOSLAMGO}.}
\label{fig1:compare GO SLAM}
\vspace{-2mm}
\end{figure}
To address the above problems, we propose MoD-SLAM, the first monocular NeRF-based dense mapping method that allows 3D reconstruction in real-time in unbounded scenes. In MoD-SLAM, we introduce a novel reparameterization approach. This method contracts space in spherical coordinates, thus eliminating the NDC \cite{Mildenhall2020NeRFRS} forward-facing scene limitations. Furthermore, unlike the sampling approach of previous NeRF-based SLAM systems, we adopt Gaussian encoding to sample information by projecting a cone towards the pixel center. This method allows for a more accurate and comprehensive capture of spatial information. To improve the accuracy of reconstruction in unbounded scenes, we further incorporate a monocular depth estimation module and a depth distillation module at the front-end. Compared to previous monocular systems \cite{Rosinol2022NeRFSLAMRD, Zhang2023HISLAMMR} that directly invert depth from the Hessian matrix, our pre-trained depth model is allowed to fine-tuned on each scene. This crucial improvement enable MoD-SLAM achieves excellent predictive performance in various scenes, a feat unprecedented in prior work. Additionally, given the challenge of pose estimation in monocular SLAM systems, particularly in unbounded scenes, we introduce a depth-supervised camera tracking method. More specifically, we add a robust depth loss term to constrain scales, enabling the system to estimate poses more accurately. Extensive evaluations on a wide range of indoor RGB sequences demonstrate the scalability and predictive power of our approach. Overall, we make the following contributions:\newline
- We propose the first monocular NeRF-based dense SLAM system that allow unbounded scene representation and achieve high-quality novel view synthesis by estimating the depth values prior.\\
- We introduce a ray reparameterization strategy and a Gaussian encoding input that contract the scene and capture spatial features more accurately to efficiently represent unbounded scenes.\\
- We introduce a depth-supervised camera tracking method to improve camera pose estimation in monocular SLAM. By combining loop closure detection and global optimization, we demonstrate state-of-the-art performance in both mapping and tracking metrics of our system.

\section{Related work}

\textbf{Monocular Visual SLAM}\quad In the realm of robotic navigation and augmented reality, Simultaneous Localization and Mapping (SLAM) plays a pivotal role. Monocular Visual SLAM has attracted significant attention due to its cost-effectiveness compared to stereo SLAM. This technique constructs the 3D map of the surroundings while simultaneously estimating the camera's position and orientation within that map. By processing sequential frames of video, monocular SLAM algorithms detect and track visual features across images, using them to calculate depth and structure from motion. Some works focused on using more efficient data structures like VoxelHashing \cite{10.1145/2461912.2461940,Stotko2018SLAMCastLR,10.1145/2508363.2508374} and Octrees \cite{Vespa2018EfficientOV,10.1016/j.gmod.2012.09.002} to improve performance. The most notable method in this domain is ORB-SLAM3 \cite{Campos2020ORBSLAM3AA}, a versatile and accurate system that utilizes ORB features to achieve robust tracking and mapping under a wide range of conditions. Another innovative approach is DROID-SLAM \cite{Teed2021DROIDSLAMDV}, which introduces deep learning to enhance tracking accuracy and the overall robustness of the SLAM system. While these methods have set high benchmarks, they are primarily reliant on feature extraction and matching, which can be challenging in feature-poor or dynamic environments.\\
\\
\textbf{NeRF-based SLAM}\quad Recent advances in the Neural Radiance Field have shown powerful ability in scene reconstruction. In this evolutionary sequence, NICE-SLAM \cite{Zhu2021NICESLAMNI} and iMAP \cite{Sucar2021iMAPIM} were the first systems to incorporate these concepts, effectively employing NeRF \cite{Mildenhall2020NeRFRS} to render intricate environmental geometry by optimizing neural fields. However, the lack of loop closure detection results in a susceptibility to drift over time, especially in longer sequences or larger environments where re-localization is critical. GO-SLAM \cite{Zhang2023GOSLAMGO} introduces loop closure detection based on the previous work. It determines loop closure by detecting co-visibility between the current and historical keyframes and also attempts to map by relying only on RGB inputs. It has made significant progress in solving the scale drift problem, but the lack of the original depth information of the scene in its RGB input mode leads to a large error in scene scale reconstruction. Subsequently, HI-SLAM \cite{Zhang2023HISLAMMR} emerged, striving to overcome the flaws inherent in GO-SLAM \cite{Zhang2023GOSLAMGO} by implementing a more efficient loop closure and a priori depth estimates from Hessian matrices. However, the systems mentioned above are incapable of mapping in unbounded scenes.
\begin{figure}[t]
\includegraphics[width=122mm]{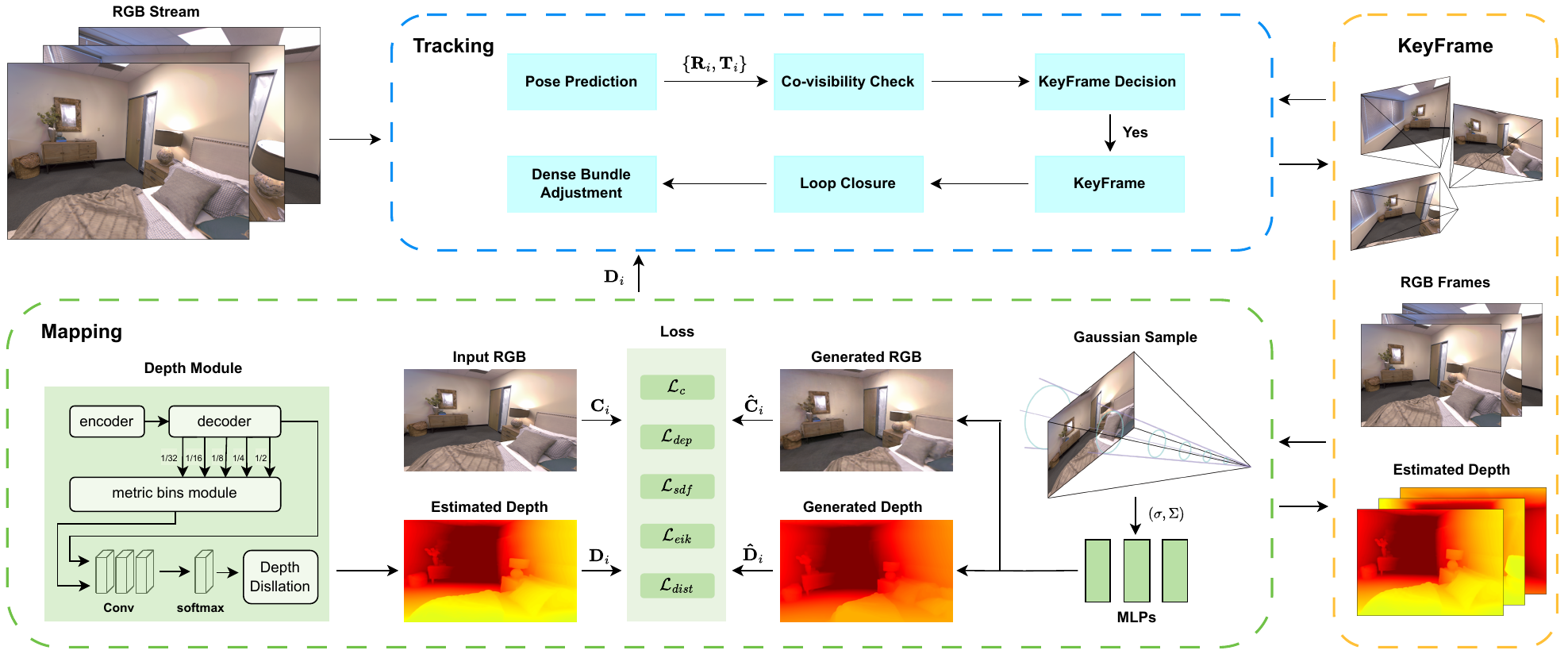}
\captionsetup{justification=justified,singlelinecheck=false}
\caption{\textbf{Overview of MoD-SLAM.} We demonstrate the monocular mode of our system, which uses RGB streams as its input. 1) Within the tracking process, the system performs depth-supervised camera pose prediction and executes loop closure and global optimization based on a co-visibility check to refine camera poses and correct drift. Simultaneously, it selects keyframes to fed into the mapping module. 2) The mapping process leverages the RGB map derived from keyframes to forecast depth values via a depth estimation and depth distillation modules. Concurrently, Gaussian encoding and contraction function are applied to the keyframes, and the resultant spatial mean and covariance are fed into MLPs to reconstruct unbounded scenes.}
\label{fig1:Overview process}
\end{figure}
\vspace{-2mm}
\section{Method}
Fig. \ref{fig1:Overview process} shows an overview of our method. We aim to estimate the camera poses $\left\{\mathbf{R}_{i}, \mathbf{T}_{i}\right\}_{i=1}^M$ of every frame and simultaneously reconstruct the unbounded scene with RGB streams input. In Sec. \ref{SectionA:scene}, we introduce the spherical contraction function to handle unbounded large scenes and use Gaussian encoding for a more precise geometric structure and visual appearance information $\left\{{\sigma}_{i}, \mathbf{\Sigma}_{i}\right\}_{i=1}^M$ capture. To address the scale inconsistency of monocular unbounded scene reconstruction, in Sec. \ref{SectionB:depth}, we design a monocular depth estimation module and depth distillation module to extract accurate depth values $\left\{\mathbf{D}_{i}\right\}_{i=1}^M$, constraining the scale in the currently observed scene. In the camera tracking process, Sec. \ref{SectionD:tracking} introduces a robust depth penalty term to significantly improve pose prediction accuracy based on depth values generated by the depth estimation module. Sec \ref{SectionE: mapping} introduces a NeRF-based volume rendering method, which uses color values $\bigl\{\hat{\mathbf{C}}_{i}\bigr\}_{i=1}^M$ and depth values $\bigl\{\hat{\mathbf{D}}_{i}\bigr\}_{i=1}^M$ for networks training.

\subsection{Gaussian Scene Representation} \label{SectionA:scene}
\textbf{Multivariate Gaussian Encoding} \quad NeRF constructs the training dataset by projecting a ray to the center of each pixel point and then performing coarse to fine sampling. This approach leads to missing information as rays only capture information along extension lines of pixel centers rather than on three-dimensional spaces. Inspired by the Mip-NeRF \cite{Barron2021MipNeRFAM}, we direct a cone-shaped projection to the pixel rather than slender rays. Projecting the cone allows us to collect more detailed information in the three-dimensional space and transfer it to MLPs for training. The training speed of the model is also greatly improved by eliminating the coarse to fine sampling process.

To compute the above features, we approximate the conic truncation with multivariate Gaussians to compute features in a particular space. We introduce $t_{\mu}=(t_{0}+t_{1})/2$ and $t_{\delta}=(t_{0}-t_{1})/2$ as refined parameters to replace neighboring sampling points $t_{0}$ and $t_{1}$ to bolster the system's stability. The multivariate Gaussian features are represented by three variables: the mean distance along the ray $\mu_{t}$, the variance along the ray $\sigma_t^2$, and the variance perpendicular to the ray $\sigma_r^2$:
\begin{equation}
\centering
\mu_{t}=t_{\mu} + \frac{2t_{\mu}t_\delta^2}{3t_\mu^2+t_\delta^2},\quad
\sigma_t^2=\frac{t_\delta^2}{3} - \frac{4t_\delta^4(12t_\mu^2-t_\delta^2)}{15(3t_\mu^2+t_\delta^2)^2},
\end{equation}
\begin{equation}
\centering
\sigma_r^2 = r^2(\frac{t_\delta^2}{4}+\frac{5t_\delta^2}{12}-\frac{4t_\delta^4}{15(3t_\mu^2+t_\delta^2)}),
\end{equation}
r is the perpendicular distance between the sampling plane where the ray is located and the ray at the center of the pixel point after transformation by spherical coordinates. Define the position of the origin of the camera in the world coordinate system as $\mu = \mathbf{o}+\mu_{t}\mathbf{d}$. To transform the mean and covariance of the variables expressing the multivariate Gaussian from the camera coordinate system to the world coordinate system, we use equations as:
\begin{equation}
\centering
\sigma = \mathbf{d}\mu_{t}, \quad 
\mathbf{\Sigma} = \mu_{r}(\mathbf{I}-\mathbf{d}\mathbf{d^T}) + \mu_{t}\mathbf{d}\mathbf{d^T},
\label{sigama}
\end{equation}
following the above equations, we can obtain a multivariate Gaussian representation for positions as well as directions in a particular space.
\begin{figure}[t]
\includegraphics[width=121mm]{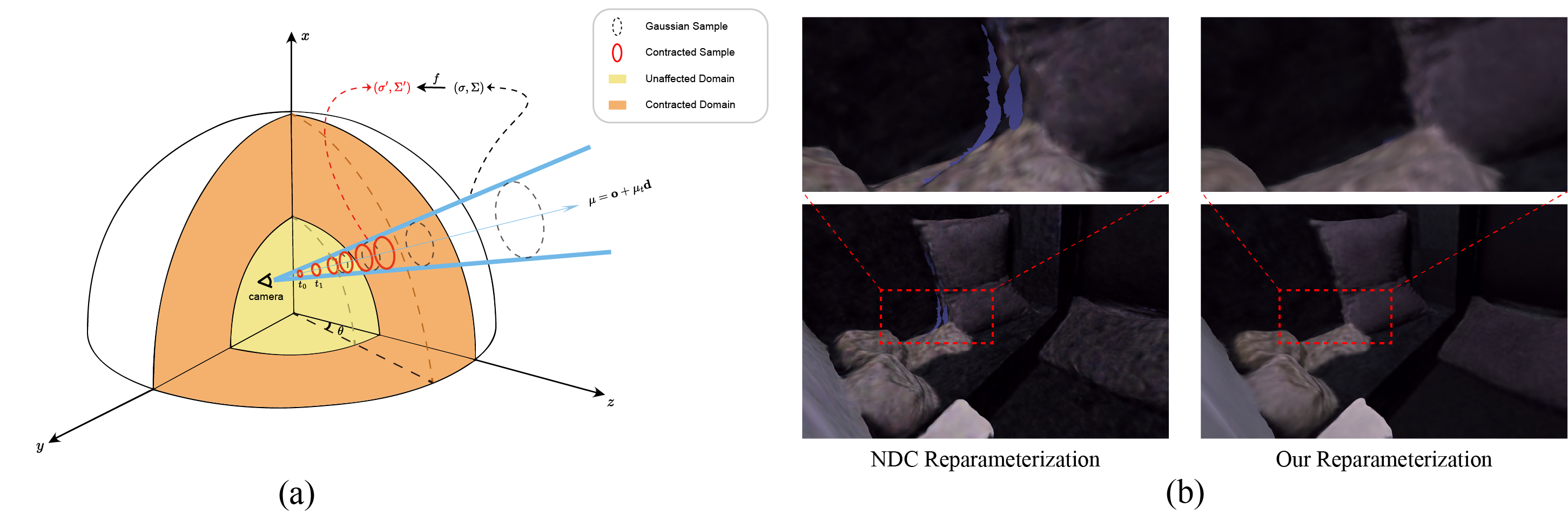}
\captionsetup{justification=justified,singlelinecheck=false}
\caption{\textbf{(a) Scene Reparameterization.} To manage unbounded scenes, we use a contract function to map mean and covariance data from spatial sampling (black dashed line) directly to a new range (red solid line). We save the data within the sphere of radius 1 (yellow region) while mapping the data from regions beyond a radius of 1 into the spherical coordinate system with a radius of 2 (orange region). \textbf{(b) NDC vs Our Reparameterization.} NDC results in voids at the boundaries of geometric objects and lacks the feature extraction and reconstruction quality achieved by our reparameterization approach.}
\label{fig2:Unbounded Scene}
\end{figure}
\\
\\
\textbf{Unbounded Scenes}\quad Previous research \cite{Barron2021MipNeRF3U} has demonstrated that the predictive efficacy of MLPs significantly diminishes when applied to datasets characterized by large disparities. For scenes without defined boundaries, normalization of the scene is impracticable owing to the lack of essential boundary scale information. Therefore, we define $f (\mathbf{x})$ to reparameterize the data as follows:
\begin{equation}
\centering
f(\mathbf{x}) = \lambda \mathbf{x} + (1-\lambda)(2-\frac{1}{\lVert \mathbf{x} \rVert})(\frac{\mathbf{x}}{\lVert \mathbf{x} \rVert}), \quad
\lambda = 
\begin{cases}
1 \quad \lVert \mathbf{x} \rVert \le 1\\
0 \quad \lVert \mathbf{x} \rVert > 1
\label{Equation:Unbound}
\end{cases}
\end{equation}
This can be interpreted as an extension of the normalized device coordinate (NDC). We make the scene shrink along the inverse depth direction, thus enabling the network to train on and interpret unbounded geometric scenes. Since in Equ. \ref{sigama} above we used the mean and covariance to represent information about the cross-section within the conic, we also need to map the mean and covariance to it. 

To achieve this, first, we linearize $f(\mathbf{x})$ by performing a first-order Taylor expansion at the point $\mathbf{\mu}$:
\begin{equation}
\centering
f(\mathbf{x})=f(\mathbf{\mu})+\mathbf{J}_{f}(\mathbf{\mu})(\mathbf{x}-\mathbf{\mu}),
\end{equation}
where $\mathbf{J}_{f}(\mathbf{\mu})$ is the jacobian of $f$ at $\mathbf{\mu}$. After doing that, we can obtain mean $E(\mathbf{x})$ and covariance $D(\mathbf{x})$ with our choice for $f$ based on Equ. \ref{Equation:Unbound}:
\begin{equation}
\centering
f(E(\mathbf{x}),D(\mathbf{x}))=
\begin{cases}
E(\mathbf{x}): f(\mathbf{\mu})\\
D(\mathbf{x}): \mathbf{J}_{f}(\mathbf{\mu})D(\mathbf{x})\mathbf{J}_{f}(\mathbf{\mu})^T.
\end{cases}
\end{equation}

Fig. \ref{fig2:Unbounded Scene}(a) presents the visualization of our reparameterization. Compared to NDC reparameterization, our method solves the issue of missing scene information caused by the misalignment of the camera viewpoint with the primary object. Fig. \ref{fig2:Unbounded Scene}(b) demonstrates that our method achieves better hole-filling result at the boundaries of geometric objects than NDC.

\subsection{Prior Depth Estimation}
\label{SectionB:depth}
The absence of a predefined scale in unbounded scenes poses a significant challenge to monocular reconstruction, leading to pronounced scale disparity. To improve the accuracy of our monocular implicit neural representation in unbounded scenes, we incorporate prior depth estimation for scale constraints. Our proposed monocular SLAM system contains a depth estimation module and a depth distillation module. More precisely, the input RGB stream serves dual purposes: first, as a dataset for normal tracking and mapping process, and second, as input to a pre-trained depth estimation module. This module generates depth information for the scene, which is then distilled through a depth distillation module to refine the depth value. With fine-tuning pre-trained depth estimation module, compared to previous monocular depth estimation methods, our method achieve more accurate depth estimation in various scenes.
\\
\\
\textbf{Depth Estimation}\quad Fig. \ref{fig1:Overview process} shows the components of our depth estimation module. Our depth estimation module contains two parts: relative depth estimation and metric depth estimation. In the relative depth estimation module, we adopt the training criterion of MiDaS \cite{Ranftl2019TowardsRM} and use the DPT decoder-encoder architecture \cite{Ranftl2021VisionTF} as backbones for feature extraction to predict the relative depths. In the metric depth estimation module, we follow ZoeDepth \cite{Bhat2023ZoeDepthZT} by proposing the metric bins module to output the metric depth.

The findings from \cite{Bhat2023ZoeDepthZT,Li2023PatchFusionAE} indicate that utilizing a singular metric head across various datasets fails to yield effective predictions. Moreover, we find that predicting depth using only the pre-trained ZoeDepth model leads to scale non-conformity. Therefore, we set multiple metric heads in our model and fine-tune ZoeDepth using specific datasets. These strategies constrain the scale of the depth values predicted by our depth estimation module. \\
\\
\textbf{Depth Distillation}\quad Due to constraints within the dataset, coarse depth annotations, the depth information derived from single-view depth estimation may deviate from actual depth data. To address this issue, we introduce a depth distillation module. Building on the foundation laid by prior research \cite{Wang2023SparseNeRFDD,Niemeyer2021RegNeRFRN}, our module focuses on two key aspects of depth map: spatial continuity and spatial correspondence.

Given a pixel set $\mathbf{P}$ of an RGB image $\mathbf{I}$, we calculate the depth $\mathbf{\widehat{D}}_{k}$ of each pixel point $k_{n}\in\mathbf{P}$. Simultaneously, we extract the depth $\mathbf{D}_{k}$ which shares the same spatial position as $\mathbf{\widehat{D}}_{k}$ from the depth estimation module. We ensure the spatial correspondence of the depth map by introducing a spatial correspondence regularization term. We randomly sample two pixels $k_{m}, k_{n}\in\mathbf{P}$, the depth spatial correspondence loss is introduced as:
\begin{equation}
\centering
\mathcal{L}_{cor} = \frac{1}{M}\sum_{k_{m}, k_{n} \in \mathbf{P}}\sum_{\mathbf{D}_{k_{m}}\le\mathbf{D}_{k_{n}}}\mathbf{max}(\widehat{\mathbf{D}}_{k_{m}}-\widehat{\mathbf{D}}_{k_{n}}+\tau,0),
\label{Spatial cor. loss}
\end{equation}
where $M$ represents the count of $\mathbf{D}_{k_{m}}\le\mathbf{D}_{k_{n}}$ is observed among the total samples taken. \(\tau\) is a minimal constant introduced to accommodate slight variations in the depth map. 

Furthermore, we propose the spatial continuity regularization term to constrain spatial continuity between two depth maps. To achieve it, we randomly sample pixel $k_{m}\in \mathbf{P}$. Then, within $k_{m}$ neighboring 3x3 pixel grids, we extract the depth value of each grid denoted as the set $\mathbf{\Theta}_{k_{m}}$. The depth spatial continuity loss is given by:
\begin{equation}
\centering
\mathcal{L}_{con} = \frac{1}{M}\sum_{k_{m}: T(k_{m})\ge4}\sum_{\mathbf{D}_{k_{n}}\in \mathbf{\Theta_{k_{m}}}}\mathbf{max}(|\widehat{\mathbf{D}}_{k_{m}}-\widehat{\mathbf{D}}_{k_{n}}|+\tau',0),
\label{Spatial con. loss}
\end{equation}
where $T(k_{m})$ represents the count of $\mathbf{D}_{k_{n}}\in \mathbf{\Theta}_{k_{m}}$, holding $|\mathbf{D}_{k_{m}}-\mathbf{D}_{k_{n}}|< \tau'$ true. $\tau'$ signifies a minor margin between the current pixel point and its adjacent pixel points. $M$ denotes the aggregate count of samples satisfying the criteria within set $\mathbf{P}$.

\subsection{Depth-supervised Camera Tracking}
\label{SectionD:tracking}
For more accurate position tracking, in this work, we employ DROID-SLAM \cite{Teed2021DROIDSLAMDV} as the front-end of our system. In traditional DROID-SLAM monocular mode of tracking, depth values are obtained by reverse-solving the Hessian matrix from the predicted poses of the GRU layer. The cost function as follow:
\begin{equation}
\centering
\mathbf{E}(\mathbf{G'},\mathbf{D'}) = \sum_{(i,j) \in \epsilon}\Vert \mathbf{p}_{ij}^* - \delta(\mathbf{p}_{i},\mathbf{d}'_{i})\Vert_{diag(w_{ij})}^2,
\end{equation}
where $\Vert\cdot\Vert_{diag()}$ is the Mahalanobis distance which weights the error terms based on the confidence weights $w_{ij}$. $\mathbf{p}_{ij}^*$ represents the corrected coordinates of pixel $\mathbf{p}_{i}$ mapped into frame $j$ and $\delta(\cdot)$ denotes the coordinates of pixel $\mathbf{p}_{i}$ mappend into frame $j$ using the refined depth. However, the depth values are subject to scale uncertainty because of the error in the pose prediction. To address this issue, we constrain depth values by introducing a new depth penalty term into the original loss function as follow:
\begin{equation}
\centering
R = \sum_{i \in \epsilon}\begin{cases} (\mathbf{d}_{i}-\mathbf{d}'_{i})^2& for 
|\mathbf{d}_{i}-\mathbf{d}'_{i}|<\tau_{tra}\\\tau_{tra}|\mathbf{d}_{i}-\mathbf{d}'_{i}|& otherwise\end{cases},
\end{equation}
where $\mathbf{d}_{i}$ denotes the depth value obtained from the depth estimation module (Sec. \ref{SectionB:depth}) and $\mathbf{d}'_{i}$ represents the depth value from Hessian matrix. $\tau_{tra}$ is the hyperparameter. Due to the error of the depth values produced by the depth estimation module, we utilize the L1 loss for outlier pixels, while employing the L2 loss for the rest. Compared to solely employing the L2 loss, our penalty term is more robust to outliers. Then, following classical optical flow-based SLAM methods \cite{Zhang2020FlowFusionDD}, we use the Gauss-Newton method to solve the Hessian matrix for obtaining updates of inverse depth and camera poses:
\begin{equation}
\left[
\begin{array}{cc}
C & E \\
E^T & P \\
\end{array}
\right]
\left[
\begin{array}{c}
\mathbf{\Delta} \mathbf{\xi} \\
\mathbf{\Delta} \mathbf{d} \\
\end{array}
\right]
=
\left[
\begin{array}{c}
\mathbf{v} \\
\mathbf{w} \\
\end{array}
\right]
\qquad
\begin{aligned}
\mathbf{\Delta \xi} &= [C - EP^{-1}E^T]^{-1}(\mathbf{v} - EP^{-1}\mathbf{w}) \\
\mathbf{\Delta d} &= P^{-1}(w - E^T \mathbf{\Delta \xi})
\end{aligned}
\end{equation}
where $C$ is the block camera matrix, and $P$ is the diagonal matrix corresponding to the points. $\mathbf{\Delta \xi}$ and $\mathbf{\Delta d}$ denotes pose and depth update values. Benefiting from prior depth values, our system can provide a more precise estimate of the pose. We also provide details on loop closure detection in Appendix.

\subsection{Mapping}
\label{SectionE: mapping}

\textbf{Depth and Color Rendering}\quad Given the current camera pose and camera intrinsic matrix, we can calculate the viewing direction \textbf{r} and change it to the world coordinate. Unlike NeRF, we propose here a more efficient sampling technique, the differential sampling technique. To accomplish this idea, we first sample the ray at the proximal $n$ as well as at the distal $f$, and then obtain a series of sampling points by sampling uniformly in this space $\{s_{1},...,s_{N}\}$. After that, we map the sampling points to the following function:
\begin{equation}
\centering
s'=\frac{1}{s\frac{1}{f}+(1-s)\frac{1}{n}},
\end{equation}
and then we get the new sampling points as $\{s_{1}',...,s_{N}'\}$.

To complete the scene reconstruction, we randomly extract $N$ rays and $M$ pixel points in the current frame and key frames. We input the position set $\mathbf{P}$ to the depth prediction module in Sec. \ref{SectionB:depth} to get the true depth $\mathbf{D}$. We do a multivariate Gaussian variation of the pose set $\left\{\mathbf{R}, \mathbf{T}\right\}$ in Sec. \ref{SectionA:scene} to get the mean as well as the covariance set. The mean and covariance are then passing into the MLPs to obtain the predicted volume density as well as the predicted color. Volume density is used to calculate the occupancy probability $o_{i}$ and the ray termination probability at each sample can then be calculated as $w_{i}=o_{i}\prod_{i=j}^{i-1} (1-o_{j})$. Eventually, the depth and color can be rendered as:
\begin{equation}
\centering
\widehat {\mathbf{D}} = \sum_{k=1}^N w_{i}d_{i}, \quad
\widehat {\mathbf{C}} = \sum_{k=1}^N w_{i}c_{i},
\end{equation}
where $d_{i}$ is the corresponding depth values of sampling points' gaussian areas and $c_{i}$ is the RGB values of sampling points' gaussian areas. \\
\\
\textbf{Loss}\quad To enhance the fidelity of the reconstructed scene to the actual scene, we initially sample $M$ pixels from the keyframes. Subsequently, the network can be optimized by minimizing the loss.

We use the method mentioned in \cite{Mller2022InstantNG} to render the color of the pixel point $\widehat{\mathbf{C}}$. To optimize the network, we minimize the loss between the predicted color and the true color:
\begin{equation}
\centering
\mathcal{L}_{rgb} = \frac{1}{M}\sum_{m=1}^M |\widehat{\mathbf{C}}_{m}-\mathbf{C}_{m}|,
\end{equation}
We also use Equ. \ref{Spatial cor. loss} and Equ. \ref{Spatial con. loss} to constrain depth value to supervise the reconstruction of this system and better recover the scale:
\begin{equation}
\centering
\mathcal{L}_{depth} = \mathcal{L}_{cor}+\lambda_{con}\mathcal{L}_{con},
\end{equation}
where $\lambda_{con}$ denotes the hyperparameter employed to modulate the weight of $\mathcal{L}_{depth}$. Furthermore, following \cite{Barron2021MipNeRF3U}, we also introduce distortion loss, which is used to eliminate artifacts in pictures caused by discontinuities in volume density over tiny areas. Distortion loss is defined as:
\begin{equation}
\centering
\mathcal{L}_{dist} = \sum_{i,j} w_{i}w_{j}\lvert\frac{s_{i}+s_{i+1}}{2} - \frac{s_{j}+s_{j+1}}{2}\rvert + \sum_{i}\frac{1}{3}w_{i}^2(s_{i+1}-s_{i}),
\end{equation}
where the first term minimizes the weighted distance between pairs of points in all intervals and the second term minimizes the weighted size of each individual interval. Additionally, following \cite{Gropp2020ImplicitGR}, regularization to the predicted SDF is also introduced in this system. At each iteration, we can get the predicted signed distance of given batch points as: $\widehat \xi=f(s;\theta)$. To encourage valid SDF values, especially in unsupervised areas, we adopt the Eikonal term:
\begin{equation}
\centering
\mathcal{L}_{eik} = \frac{1}{MN_{ray}}\sum_{m=1}^M \sum_{i\in N} (1-\lVert \bigtriangledown \phi(\mathbf{s}_{i})\rVert)^2.
\end{equation}
For the SDF, we approximate the SDF values by calculating the depth values along the observation direction as: $\mathbf{b(s)}= \widehat {\mathbf{D}} - \mathbf{D}$. We then define the adopted loss function by its distance from the surface. For $|\mathbf{\phi(s)}|\leq|\mathbf{b(s)}|$,$\forall \mathbf{s}$, the near-surface points($\widehat {\mathbf{D}} - \mathbf{D} \leq \zeta$), following \cite{Ortiz2022iSDFRN,Wang2022GOSurfNF} the SDF loss is:
\begin{equation}
\centering
\mathcal{L}_{sdf} = \frac{1}{MN_{ray}}\sum_{m=1}^M \sum_{i\in N} \lvert \phi(\mathbf{s}_{i}) - \mathbf{b}(\mathbf{s}_{i}) \rvert,
\end{equation}
for the other situation, the far-surface points($\widehat{\mathbf{D}} - \mathbf{D} > \zeta$), the SDF loss is:
\begin{equation}
\centering
\mathcal{L}_{sdf} = \frac{1}{MN_{ray}}\sum_{m=1}^M \sum_{i\in N} max(0,e^{-\alpha \mathbf{\phi}({\mathbf{s}_{i})}},\mathbf{\phi}({\mathbf{s}_{i})}-\mathbf{b}({\mathbf{s}_{i})}),
\end{equation}
where we set $\alpha$ equal to 5 in this system. 

Our real-time mapping thread optimizes the predicted neural networks by continuously feeding keyframes to achieve ideal 3D reconstruction results. The total loss of the system is:
\begin{equation}
\centering
\mathcal{L}_{total} = \lambda_{c}\mathcal{L}_{c}+\lambda_{dep}\mathcal{L}_{dep}+\lambda_{dist}\mathcal{L}_{dist} + \lambda_{eik}\mathcal{L}_{eik} + \lambda_{sdf}\mathcal{L}_{sdf},
\end{equation}
where $\lambda_{c}$, $\lambda_{depth}$, $\lambda_{dist}$, $\lambda_{eik}$, $\lambda_{sdf}$ are the hyperparameters for each loss, controlling the weight of each loss in the total loss.
\begin{figure}[t]
\centering
\includegraphics[width=120mm]{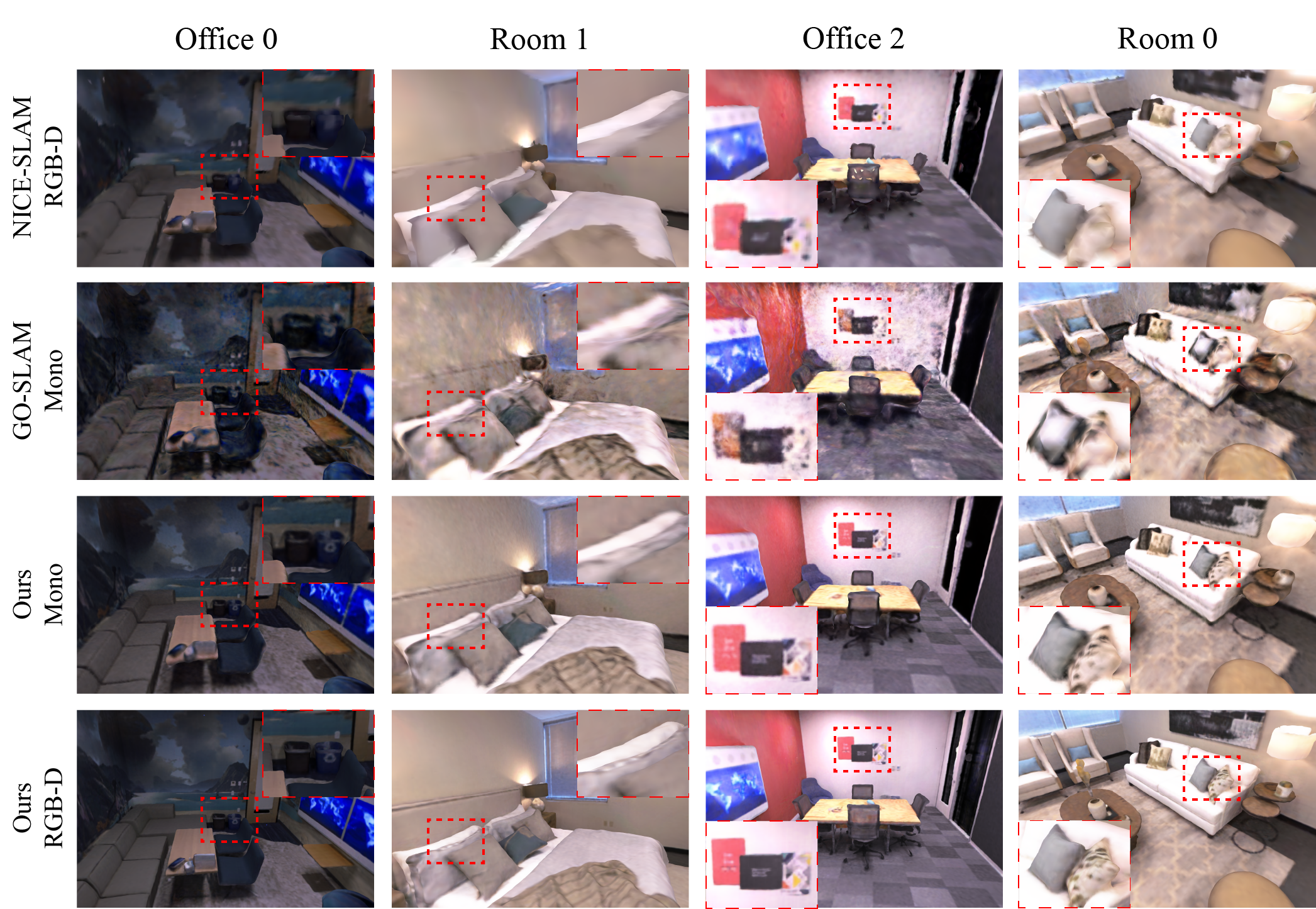}
\captionsetup{justification=justified,singlelinecheck=false}
\caption{\textbf{Reconstruction results on Replica \cite{Straub2019TheRD}}. Compared with baselines, the 3D scenes reconstructed by our method show enhanced geometric structure and texture features.}
\label{fig2:Replica}
\vspace{-4mm}
\end{figure}
\begin{figure}[t]
\centering
\includegraphics[width=120mm]{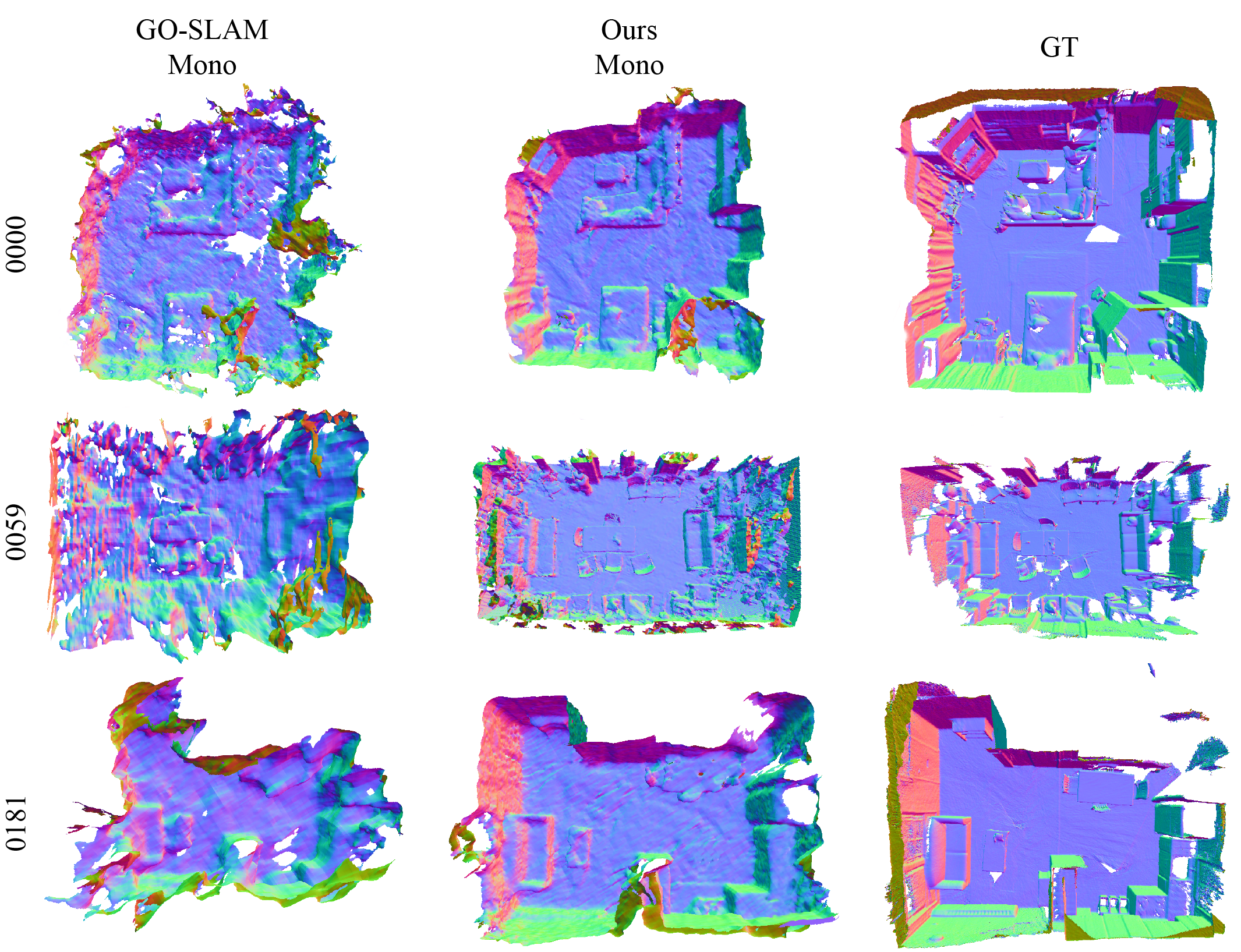}
\captionsetup{justification=justified,singlelinecheck=false}
\caption{\textbf{Reconstruction results on ScanNet \cite{Dai2017ScanNetR3}}. We show the full view of the reconstructed 3D scenes. Our system enhances the precision in scaling the scenes.}
\label{fig7:scannet_bound}
\vspace{-2mm}
\end{figure}

\section{Experiments}
\label{sec:blind}

We evaluate our models in both synthetic and real-world datasets, including Replica \cite{Straub2019TheRD} and ScanNet \cite{Dai2017ScanNetR3}. We conducted a comparative analysis of the reconstruction quality of our proposed system against state-of-the-art SLAM systems. We also show more experimental results in the Appendix.

\subsection{Experimental Setup}

\begin{figure}[t]
\centering
\includegraphics[width=120mm]{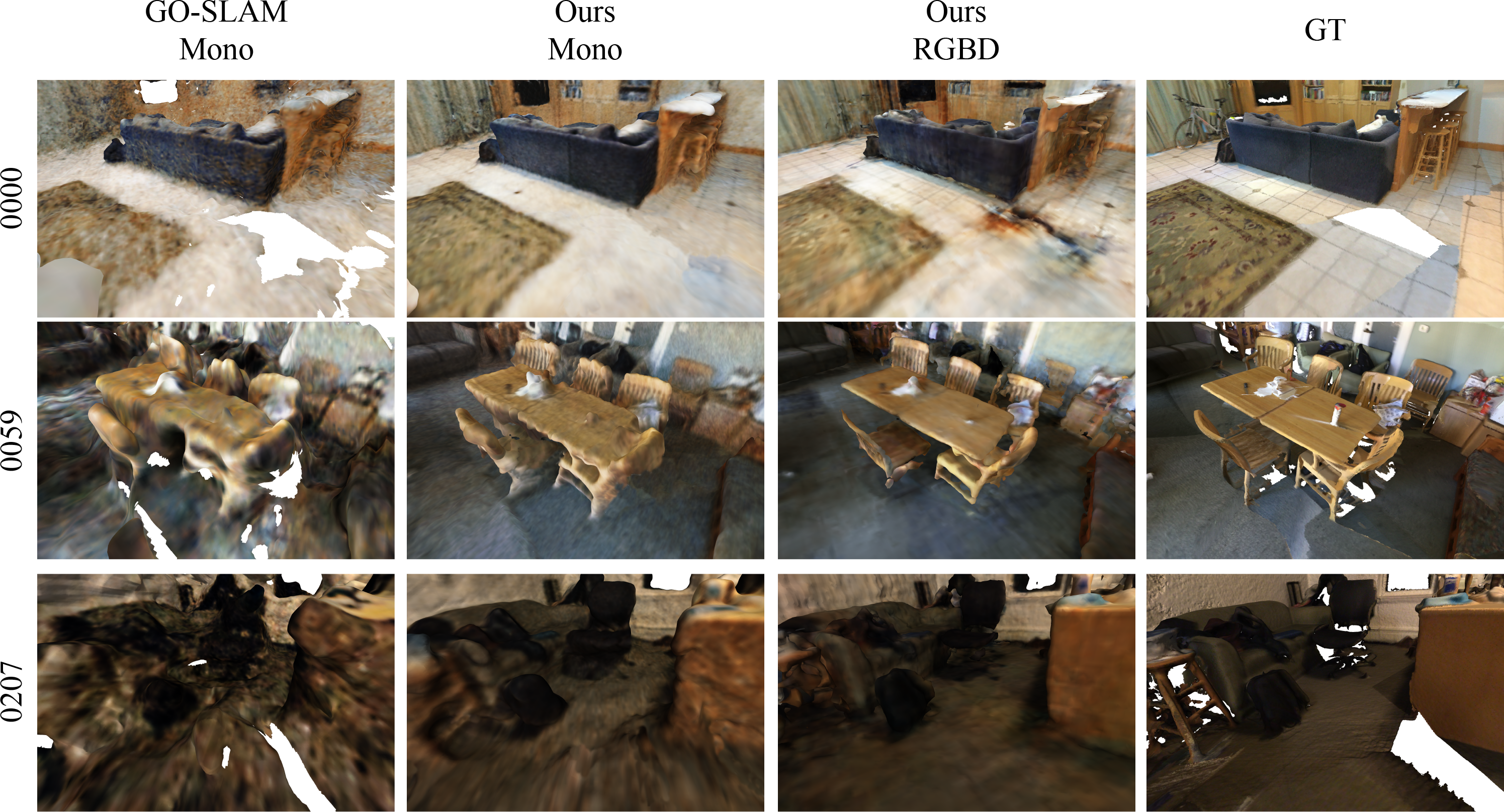}
\captionsetup{justification=justified,singlelinecheck=false}
    \caption{\textbf{Reconstruction results on ScanNet \cite{Dai2017ScanNetR3}.} Compared with baselines, our system demonstrates enhanced capability in scale reconstruction and void interpolation within the scenes. }
\label{fig3:Scannet}
\vspace{-2mm}
\end{figure}

\textbf{Datasets}\quad For our quantitative analysis, we implement our system on the Replica dataset \cite{Straub2019TheRD} and ScanNet dataset \cite{Dai2017ScanNetR3}. We follow the same pre-processing step for datasets as in NICE-SLAM \cite{Zhu2021NICESLAMNI}. Each dataset presents challenging camera dynamics, effectively showcasing our model's performance and accuracy. We concurrently utilize these datasets to assess our model in both monocular RGB and RGB-D.
\\
\\
\textbf{Metrics}\quad In evaluating camera tracking accuracy, we follow standard SLAM metrics. This involves comparing the trajectory of keyframes against the true trajectory and computing the Root Mean Square Error (RMSE) of the Absolute Trajectory Error (ATE) \cite{Sturm2012ABF}. Regarding reconstruction quality, we evaluate it by accuracy [cm], PSNR \cite{Fardo2016AFE}, SSIM and F-score [$<$ 5cm \%] \cite{Sucar2021iMAPIM}. In addition, we evaluate the L1 loss \cite{Zhu2021NICESLAMNI} between the reconstructed and real meshes. All our experimental results are averaged after three runs.
\\
\\
\textbf{Implementation Details}\quad We run our system on a PC with a 4.70 GHz AMD 7900X CPU and NVIDIA RTX 4090 GPU. We use the pre-trained model from DROID-SLAM \cite{Teed2021DROIDSLAMDV} for tracking and use the same parameters in the tracking process as it. In the reconstruction process, we choose to sample a ray at several points of $N_{strat}=20$ and $N_{imp}=40$. For loss weights, we choose $\lambda_{c}=1.0$, $\lambda_{dep}=1.0$, $\lambda_{dist}=0.002$, $\lambda_{eik}=0.15$, $\lambda_{sdf}=1.0$. The compared baseline includes classical SLAM systems \cite{Campos2020ORBSLAM3AA, Sucar2021iMAPIM}, newest SLAM systems \cite{Zhu2021NICESLAMNI, Zhang2023GOSLAMGO, Zhu2023NICERSLAMNI, Wang2023CoSLAMJC, Johari2022ESLAMED} and 3D Gaussian-based SLAM system \cite{Huang2023PhotoSLAMRS}.
\begin{table}[!t]
\caption{\textbf{Reconstruction quality and run-time comparison on Replica \cite{Straub2019TheRD}}. Our results were obtained by averaging three runs across seven scenes from the Replica dataset. '\XSolidBrush' denotes data that has yet to be released.}
\vspace{-2mm}
\centering
\scriptsize
\renewcommand{\arraystretch}{1.5}
\setlength{\tabcolsep}{1.95mm}
\begin{tabular}{l l l l l l l l}
\hline\noalign{\smallskip}
Mode &  Method & \makecell{ATE[cm]$\downarrow$} & \makecell{L1[cm]$\downarrow$} & \makecell{Acc.[cm]$\downarrow$} &  \makecell{PSNR$\uparrow$} & \makecell{SSIM$\uparrow$}& \makecell{FPS$\uparrow$}\\
\hline\noalign{\smallskip}
 & NICE-SLAM\cite{Zhu2021NICESLAMNI} & \makecell{$1.95$} & \makecell{$3.53$} & \makecell{$2.85$} & \makecell{$26.31$} & \makecell{0.847} & \makecell{$0.3$}\\
\cdashline{2-8}[0.3pt/5pt]
  RGB-D & GO-SLAM\cite{Zhang2023GOSLAMGO} & \makecell{0.34}& \makecell{3.38}& \makecell{2.50}& \makecell{27.38}& \makecell{0.851} & \makecell{\pmb{8}}\\
\cdashline{2-8}[0.3pt/5pt]
 & \textbf{Ours} & \makecell{\pmb{0.33}}& \makecell{\pmb{3.11}}& \makecell{\pmb{2.13}}& \makecell{\pmb{30.95}}& \makecell{\pmb{0.882}} & \makecell{6}\\
 \hline\noalign{\smallskip}
 & DROID-SLAM\cite{Teed2021DROIDSLAMDV} & \makecell{0.42} & \makecell{17.59} & \makecell{5.03}& \makecell{7.68}& \makecell{0.191} & \makecell{21}\\
\cdashline{2-8}[0.3pt/5pt]
 & NICER-SLAM\cite{Zhu2023NICERSLAMNI} & \makecell{1.88} &  \makecell{\XSolidBrush} & \makecell{3.65} & \makecell{25.41} & \makecell{0.827} & \makecell{6}\\
\cdashline{2-8}[0.3pt/5pt]
Mono & GO-SLAM\cite{Zhang2023GOSLAMGO} & \makecell{0.39} &  \makecell{4.39} & \makecell{3.81} & \makecell{22.13} & \makecell{0.733} & \makecell{8}\\
\cdashline{2-8}[0.3pt/5pt]
 & Photo-SLAM\cite{Huang2023PhotoSLAMRS} & \makecell{1.09} & \makecell{\XSolidBrush} & \makecell{\XSolidBrush} & \makecell{\pmb{33.30}} & \makecell{\pmb{0.930}} & \makecell{\pmb{42}}\\
\cdashline{2-8}[0.3pt/5pt]
 & \textbf{Ours} & \makecell{\pmb{0.35}} & \makecell{\pmb{3.23}} & \makecell{\pmb{2.48}} & \makecell{28.37} & \makecell{0.850} & \makecell{8}\\
\hline
\noalign{\smallskip}
\end{tabular}
\label{table:headings}
\end{table}
\begin{table}[!t]
\caption{\textbf{ATE RMSE [cm] on ScanNet \cite{Dai2017ScanNetR3}}. The performance metrics reported by our system for each scene are derived from the average of three executions.}
\vspace{-5mm}
\begin{center}
\scriptsize
\renewcommand{\arraystretch}{1.5}
\setlength{\tabcolsep}{2.0mm}
\begin{tabular}{l l l l l l l l l l}
\hline\noalign{\smallskip}
 &  Scene ID & \makecell{0000} & \makecell{0059} & \makecell{0106} &  \makecell{0169} & \makecell{0181}& \makecell{0207}& \makecell{0233}& \makecell{\pmb{Avg.}}\\
\hline
\noalign{\smallskip}
 & DROID-SLAM\cite{Zhu2021NICESLAMNI} & \makecell{$5.48$} & \makecell{$9.00$} & \makecell{\pmb{$6.76$}} & \makecell{$7.86$} & \makecell{7.41} & \makecell{$6.82$}& \makecell{$72.23$}& \makecell{$16.50$}\\
 \cdashline{2-10}[0.3pt/5pt]
  Mono & GO-SLAM\cite{Zhang2023GOSLAMGO} & \makecell{5.94}& \makecell{8.27}& \makecell{8.07}& \makecell{8.42}& \makecell{8.29} & \makecell{6.72}& \makecell{5.31}& \makecell{7.28}\\
  \cdashline{2-10}[0.3pt/5pt]
 & \textbf{Ours} & \makecell{\pmb{5.39}}& \makecell{\pmb{7.78}}& \makecell{7.64}& \makecell{\pmb{6.79}}& \makecell{\pmb{6.58}} & \makecell{\pmb{5.63}}& \makecell{\pmb{4.37}}& \makecell{\pmb{6.32}}\\
 \hline\noalign{\smallskip}
 & NICE-SLAM\cite{Zhu2021NICESLAMNI} & \makecell{8.64} & \makecell{12.25} & \makecell{8.09}& \makecell{10.28}& \makecell{12.93} & \makecell{5.59}& \makecell{7.31}& \makecell{9.29}\\
 \cdashline{2-10}[0.3pt/5pt]
 & GO-SLAM\cite{Zhang2023GOSLAMGO} & \makecell{5.35} &  \makecell{7.52} & \makecell{7.03} & \makecell{7.74} & \makecell{6.84} & \makecell{\pmb{5.26}}& \makecell{4.78}& \makecell{6.36}\\
 \cdashline{2-10}[0.3pt/5pt]
RGB-D & Co-SLAM\cite{Wang2023CoSLAMJC} & \makecell{7.13} &  \makecell{11.14} & \makecell{9.36} & \makecell{\pmb{5.90}} & \makecell{11.81} & \makecell{7.14}& \makecell{6.77}& \makecell{8.47}\\
\cdashline{2-10}[0.3pt/5pt]
 & E-SLAM\cite{Johari2022ESLAMED} & \makecell{7.32} & \makecell{8.57} & \makecell{7.53} & \makecell{6.43} & \makecell{8.85} & \makecell{5.74}& \makecell{5.13}& \makecell{7.08}\\
 \cdashline{2-10}[0.3pt/5pt]
 & \textbf{Ours} & \makecell{\pmb{5.27}} & \makecell{\pmb{7.44}} & \makecell{\pmb{6.73}} & \makecell{6.48} & \makecell{\pmb{6.14}} & \makecell{5.31}& \makecell{\pmb{3.29}}& \makecell{\pmb{5.81}}\\
\hline
\noalign{\smallskip}
\end{tabular}
\label{table1:headings}
\end{center}
\vspace{-4mm}
\end{table}
\begin{figure}[t]
\centering
\includegraphics[width=120mm]{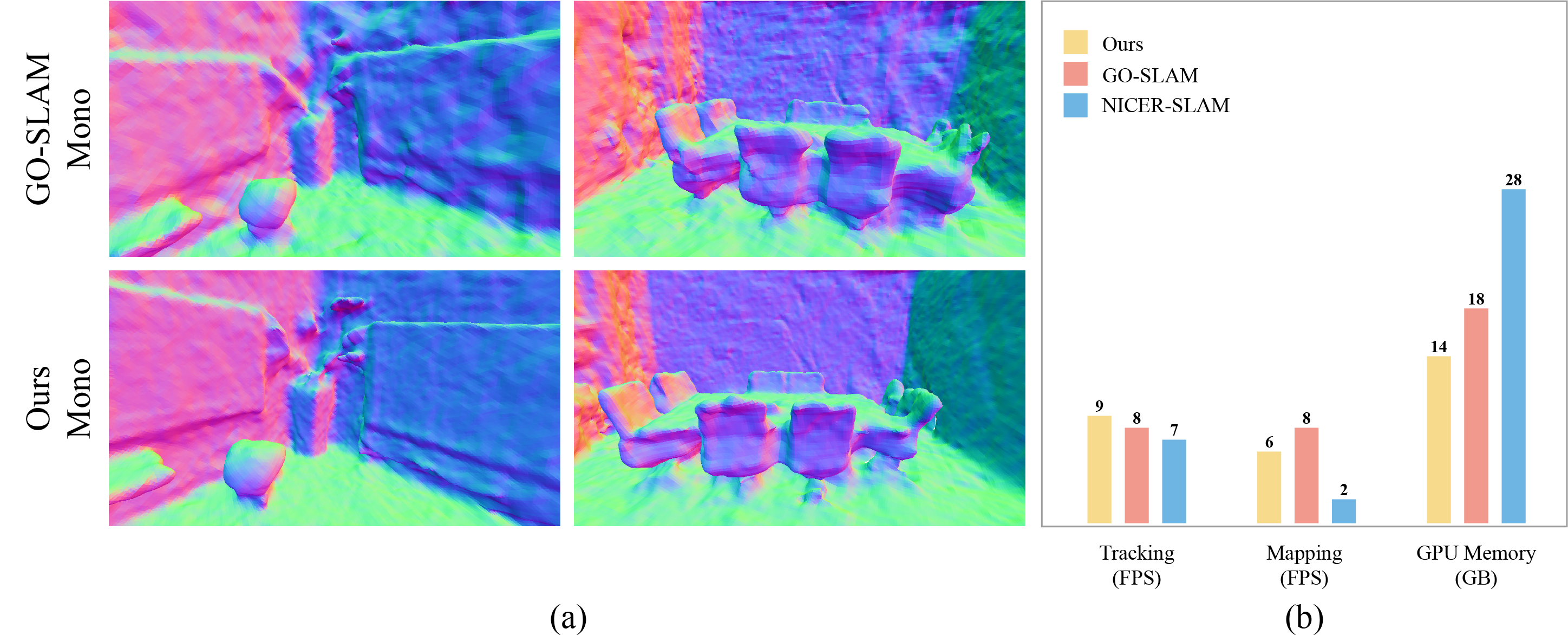}
\captionsetup{justification=justified,singlelinecheck=false, skip=5pt}
    \caption{(a) shows the reconstruction result on Replica \cite{Straub2019TheRD}. (b) shows the run time and GPU memory used in SLAM systems.}
\label{fig4:runtime}
\vspace{-5mm}
\end{figure}
\subsection{Results and Evaluation}
\textbf{Evaluation on Replica \cite{Straub2019TheRD}}\quad Table \ref{table:headings} shows the results of the reconstruction of systems on Replica. Compared to the systems mentioned in the table, our system exhibits superior tracking capabilities in both monocular and RGB-D modes. In comparison with NeRF-based SLAM  \cite{Zhu2021NICESLAMNI,Zhang2023GOSLAMGO,Zhu2023NICERSLAMNI}, our approach achieves enhanced reconstruction fidelity. Compared to 3D Gaussian-based SLAM systems \cite{Huang2023PhotoSLAMRS}, our system falls in terms of scene reconstruction accuracy. However, our system outperforms in tracking performance and is more effective in recovering the scale of the scenes. Fig. \ref{fig2:Replica} and Fig. \ref{fig4:runtime}(a) present a comparative analysis of scene reconstructions by our system against NICE-SLAM \cite{Zhu2021NICESLAMNI} and GO-SLAM \cite{Zhang2023GOSLAMGO}. Our method demonstrates superior performance.\\
\\
\textbf{Evaluation on ScanNet \cite{Dai2017ScanNetR3}}\quad We evaluated our system across seven distinct scenes in ScanNet dataset \cite{Dai2017ScanNetR3}. Table \ref{table1:headings} presents the tracking results. Our system outperforms others in tracking accuracy, particularly in the large indoor scenes of ScanNet. As illustrated in Fig. \ref{fig3:Scannet} and Fig. \ref{fig7:scannet_bound}, our system outmatches GO-SLAM \cite{Zhang2023GOSLAMGO} by rendering the scene with enhanced detail fidelity, more accurately approximating the real-world geometry and ensuring smoother scene transitions. Furthermore, our system possesses a pronounced proficiency in addressing scene sparsity through effective void interpolation.\\
\\
\textbf{Runtime Analysis}\quad We test the tracking and the mapping time of NICER-SLAM \cite{Zhu2023NICERSLAMNI}, GO-SLAM \cite{Zhang2023GOSLAMGO} and our system using the same number of sampling points. Fig. \ref{fig4:runtime}(b) shows that our system is faster than GO-SLAM and NICER-SLAM in the tracking process. However, because of the depth estimation module, our system is slightly slower than GO-SLAM in the mapping process, but higher than NICER-SLAM. In addition, Fig. \ref{fig4:runtime}(b) also demonstrates that our model size is smaller and uses less GPU memory. 

\section{Conclusion}
We presented MoD-SLAM, a monocular unbounded NeRF-based dense mapping method that demonstrates state-of-the-art performance. The proposed Gaussian-based unbounded scene representation approach allows our system to be capable of reconstructing unbounded scenes. The depth estimation module and depth distillation module provide prior depth values, addressing scale inconsistency in monocular unbounded scene reconstruction. The depth-supervised camera tracking method combines traditional SLAM method with the depth loss term, which achieves robust and accurate camera tracking and pose estimation. Our experiments show that MoD-SLAM has superior performance in both localization and reconstruction with low time and GPU memory consumption compared to state-of-the-art SLAM systems.

\clearpage

\bibliographystyle{splncs04}
\bibliography{main}

\begin{thebibliography}{10}
\providecommand{\url}[1]{\texttt{#1}}
\providecommand{\urlprefix}{URL }
\providecommand{\doi}[1]{https://doi.org/#1}

\bibitem{Barron2021MipNeRFAM}
Barron, J.T., Mildenhall, B., Tancik, M., Hedman, P., Martin-Brualla, R., Srinivasan, P.P.: Mip-nerf: A multiscale representation for anti-aliasing neural radiance fields. 2021 IEEE/CVF International Conference on Computer Vision (ICCV)  (2021)

\bibitem{Barron2021MipNeRF3U}
Barron, J.T., Mildenhall, B., Verbin, D., Srinivasan, P.P., Hedman, P.: Mip-nerf 360: Unbounded anti-aliased neural radiance fields. 2022 IEEE/CVF Conference on Computer Vision and Pattern Recognition (CVPR)  (2021)

\bibitem{Bhat2023ZoeDepthZT}
Bhat, S., Birkl, R., Wofk, D., Wonka, P., Muller, M.: Zoedepth: Zero-shot transfer by combining relative and metric depth. ArXiv  (2023)

\bibitem{Campos2020ORBSLAM3AA}
Campos, C., Elvira, R., Rodr'iguez, J.J.G., Montiel, J.M.M., Tard{\'o}s, J.D.: Orb-slam3: An accurate open-source library for visual, visual–inertial, and multimap slam. IEEE Transactions on Robotics  \textbf{37},  1874--1890 (2020)

\bibitem{10.1145/2461912.2461940}
Chen, J., Bautembach, D., Izadi, S.: Scalable real-time volumetric surface reconstruction. ACM Trans. Graph.  (2013)

\bibitem{Chung2022OrbeezSLAMAR}
Chung, C.M., Tseng, Y.C., Hsu, Y.C., Shi, X.Q., Hua, Y.H., Yeh, J.F., Chen, W.C., Chen, Y.T., Hsu, W.H.: Orbeez-slam: A real-time monocular visual slam with orb features and nerf-realized mapping. 2023 IEEE International Conference on Robotics and Automation (ICRA)  (2022)

\bibitem{Dai2017ScanNetR3}
Dai, A., Chang, A.X., Savva, M., Halber, M., Funkhouser, T.A., Nie{\ss}ner, M.: Scannet: Richly-annotated 3d reconstructions of indoor scenes. 2017 IEEE Conference on Computer Vision and Pattern Recognition (CVPR)  (2017)

\bibitem{Deng2021DepthsupervisedNF}
Deng, K., Liu, A., Zhu, J.Y., Ramanan, D.: Depth-supervised nerf: Fewer views and faster training for free. 2022 IEEE/CVF Conference on Computer Vision and Pattern Recognition (CVPR)  (2021)

\bibitem{Fardo2016AFE}
Fardo, F.A., Conforto, V.H., de~Oliveira, F.C., Rodrigues, P.S.S.: A formal evaluation of psnr as quality measurement parameter for image segmentation algorithms. ArXiv  (2016)

\bibitem{Gropp2020ImplicitGR}
Gropp, A., Yariv, L., Haim, N., Atzmon, M., Lipman, Y.: Implicit geometric regularization for learning shapes. ArXiv  (2020)

\bibitem{Huang2023PhotoSLAMRS}
Huang, H., Li, L., Cheng, H., Yeung, S.K.: Photo-slam: Real-time simultaneous localization and photorealistic mapping for monocular, stereo, and rgb-d cameras. ArXiv  (2023)

\bibitem{Johari2022ESLAMED}
Johari, M.M., Carta, C., Fleuret, F.: Eslam: Efficient dense slam system based on hybrid representation of signed distance fields. 2023 IEEE/CVF Conference on Computer Vision and Pattern Recognition (CVPR)  (2022)

\bibitem{Li2023PatchFusionAE}
Li, Z., Bhat, S.F., Wonka, P.: Patchfusion: An end-to-end tile-based framework for high-resolution monocular metric depth estimation. ArXiv  (2023)

\bibitem{Matsuki2023NEWTONNV}
Matsuki, H., Tateno, K., Niemeyer, M., Tombari, F.: Newton: Neural view-centric mapping for on-the-fly large-scale slam. ArXiv  (2023)

\bibitem{Mildenhall2020NeRFRS}
Mildenhall, B., Srinivasan, P.P., Tancik, M., Barron, J.T., Ramamoorthi, R., Ng, R.: Nerf: Representing scenes as neural radiance fields for view synthesis. Commun. ACM  (2020)

\bibitem{Mller2022InstantNG}
M{\"u}ller, T., Evans, A., Schied, C., Keller, A.: Instant neural graphics primitives with a multiresolution hash encoding. ACM Transactions on Graphics (TOG)  (2022)

\bibitem{MurArtal2016ORBSLAM2AO}
Mur-Artal, R., Tard{\'o}s, J.D.: Orb-slam2: An open-source slam system for monocular, stereo, and rgb-d cameras. IEEE Transactions on Robotics  (2016)

\bibitem{Niemeyer2021RegNeRFRN}
Niemeyer, M., Barron, J.T., Mildenhall, B., Sajjadi, M.S.M., Geiger, A., Radwan, N.: Regnerf: Regularizing neural radiance fields for view synthesis from sparse inputs. 2022 IEEE/CVF Conference on Computer Vision and Pattern Recognition (CVPR)  (2021)

\bibitem{Niemeyer2019DifferentiableVR}
Niemeyer, M., Mescheder, L.M., Oechsle, M., Geiger, A.: Differentiable volumetric rendering: Learning implicit 3d representations without 3d supervision. 2020 IEEE/CVF Conference on Computer Vision and Pattern Recognition (CVPR)  (2019)

\bibitem{10.1145/2508363.2508374}
Nie\ss{}ner, M., Zollh\"{o}fer, M., Izadi, S., Stamminger, M.: Real-time 3d reconstruction at scale using voxel hashing. ACM Trans. Graph.  (2013)

\bibitem{Ortiz2022iSDFRN}
Ortiz, J., Clegg, A., Dong, J., Sucar, E., Novotn{\'y}, D., Zollhoefer, M., Mukadam, M.: isdf: Real-time neural signed distance fields for robot perception. ArXiv  (2022)

\bibitem{Park2019DeepSDFLC}
Park, J.J., Florence, P.R., Straub, J., Newcombe, R.A., Lovegrove, S.: Deepsdf: Learning continuous signed distance functions for shape representation. 2019 IEEE/CVF Conference on Computer Vision and Pattern Recognition (CVPR)  (2019)

\bibitem{Ranftl2021VisionTF}
Ranftl, R., Bochkovskiy, A., Koltun, V.: Vision transformers for dense prediction. 2021 IEEE/CVF International Conference on Computer Vision (ICCV)  (2021)

\bibitem{Ranftl2019TowardsRM}
Ranftl, R., Lasinger, K., Hafner, D., Schindler, K., Koltun, V.: Towards robust monocular depth estimation: Mixing datasets for zero-shot cross-dataset transfer. IEEE Transactions on Pattern Analysis and Machine Intelligence  (2019)

\bibitem{Rosinol2022NeRFSLAMRD}
Rosinol, A., Leonard, J.J., Carlone, L.: Nerf-slam: Real-time dense monocular slam with neural radiance fields. 2023 IEEE/RSJ International Conference on Intelligent Robots and Systems (IROS)  (2022)

\bibitem{Sitzmann2020ImplicitNR}
Sitzmann, V., Martel, J.N.P., Bergman, A.W., Lindell, D.B., Wetzstein, G.: Implicit neural representations with periodic activation functions. ArXiv  (2020)

\bibitem{Stotko2018SLAMCastLR}
Stotko, P., Krumpen, S., Hullin, M.B., Weinmann, M., Klein, R.: Slamcast: Large-scale, real-time 3d reconstruction and streaming for immersive multi-client live telepresence. IEEE Transactions on Visualization and Computer Graphics  (2018)

\bibitem{Straub2019TheRD}
Straub, J., Whelan, T., Ma, L., Chen, Y., Wijmans, E., Green, S., Engel, J.J., Mur-Artal, R., Ren, C.Y., Verma, S., Clarkson, A., Yan, M., Budge, B., Yan, Y., Pan, X., Yon, J., Zou, Y., Leon, K., Carter, N., Briales, J., Gillingham, T., Mueggler, E., Pesqueira, L., Savva, M., Batra, D., Strasdat, H.M., Nardi, R.D., Goesele, M., Lovegrove, S., Newcombe, R.A.: The replica dataset: A digital replica of indoor spaces. ArXiv  (2019)

\bibitem{Sturm2012ABF}
Sturm, J., Engelhard, N., Endres, F., Burgard, W., Cremers, D.: A benchmark for the evaluation of rgb-d slam systems. 2012 IEEE/RSJ International Conference on Intelligent Robots and Systems  (2012)

\bibitem{Sucar2021iMAPIM}
Sucar, E., Liu, S., Ortiz, J., Davison, A.J.: imap: Implicit mapping and positioning in real-time. 2021 IEEE/CVF International Conference on Computer Vision (ICCV)  (2021)

\bibitem{Sucar2020NodeSLAMNO}
Sucar, E., Wada, K., Davison, A.J.: Nodeslam: Neural object descriptors for multi-view shape reconstruction. 2020 International Conference on 3D Vision (3DV)  (2020)

\bibitem{Tancik2023NerfstudioAM}
Tancik, M., Weber, E., Ng, E., Li, R., Yi, B., Kerr, J., Wang, T., Kristoffersen, A., Austin, J., Salahi, K., Ahuja, A., McAllister, D., Kanazawa, A.: Nerfstudio: A modular framework for neural radiance field development. ACM SIGGRAPH 2023 Conference Proceedings  (2023)

\bibitem{Teed2018DeepV2DVT}
Teed, Z., Deng, J.: Deepv2d: Video to depth with differentiable structure from motion. ArXiv  (2018)

\bibitem{Teed2021DROIDSLAMDV}
Teed, Z., Deng, J.: Droid-slam: Deep visual slam for monocular, stereo, and rgb-d cameras. In: Neural Information Processing Systems (2021)

\bibitem{Vespa2018EfficientOV}
Vespa, E., Nikolov, N., Grimm, M., Nardi, L., Kelly, P.H.J., Leutenegger, S.: Efficient octree-based volumetric slam supporting signed-distance and occupancy mapping. IEEE Robotics and Automation Letters  (2018)

\bibitem{Wang2023SparseNeRFDD}
Wang, G., Chen, Z., Loy, C.C., Liu, Z.: Sparsenerf: Distilling depth ranking for few-shot novel view synthesis. 2023 IEEE/CVF International Conference on Computer Vision (ICCV)  (2023)

\bibitem{Wang2023CoSLAMJC}
Wang, H., Wang, J., de~Agapito, L.: Co-slam: Joint coordinate and sparse parametric encodings for neural real-time slam. 2023 IEEE/CVF Conference on Computer Vision and Pattern Recognition (CVPR)  (2023)

\bibitem{Wang2022GOSurfNF}
Wang, J., Bleja, T., de~Agapito, L.: Go-surf: Neural feature grid optimization for fast, high-fidelity rgb-d surface reconstruction. 2022 International Conference on 3D Vision (3DV)  (2022)

\bibitem{10.1016/j.gmod.2012.09.002}
Zeng, M., Zhao, F., Zheng, J., Liu, X.: Octree-based fusion for realtime 3d reconstruction. Graph. Models  (2013)

\bibitem{Zhang2020FlowFusionDD}
Zhang, T., Zhang, H., Li, Y., Nakamura, Y., Zhang, L.: Flowfusion: Dynamic dense rgb-d slam based on optical flow. 2020 IEEE International Conference on Robotics and Automation (ICRA)  (2020)

\bibitem{Zhang2023HISLAMMR}
Zhang, W., Sun, T., Wang, S., Cheng, Q., Haala, N.: Hi-slam: Monocular real-time dense mapping with hybrid implicit fields. IEEE Robotics and Automation Letters  (2023)

\bibitem{Zhang2023GOSLAMGO}
Zhang, Y., Tosi, F., Mattoccia, S., Poggi, M.: Go-slam: Global optimization for consistent 3d instant reconstruction. 2023 IEEE/CVF International Conference on Computer Vision (ICCV)  (2023)

\bibitem{Zhi2019SceneCodeMD}
Zhi, S., Bloesch, M., Leutenegger, S., Davison, A.J.: Scenecode: Monocular dense semantic reconstruction using learned encoded scene representations. 2019 IEEE/CVF Conference on Computer Vision and Pattern Recognition (CVPR)  (2019)

\bibitem{Zhu2023NICERSLAMNI}
Zhu, Z., Peng, S., Larsson, V., Cui, Z., Oswald, M.R., Geiger, A., Pollefeys, M.: Nicer-slam: Neural implicit scene encoding for rgb slam. ArXiv  (2023)

\bibitem{Zhu2021NICESLAMNI}
Zhu, Z., Peng, S., Larsson, V., Xu, W., Bao, H., Cui, Z., Oswald, M.R., Pollefeys, M.: Nice-slam: Neural implicit scalable encoding for slam. 2022 IEEE/CVF Conference on Computer Vision and Pattern Recognition (CVPR)  (2021)

\bibitem{Zubizarreta2019DirectSM}
Zubizarreta, J.A., Aguinaga, I., Montiel, J.M.M.: Direct sparse mapping. IEEE Transactions on Robotics  \textbf{36},  1363--1370 (2019)

\end{thebibliography}
\clearpage

\appendix
\section{Appendix}
\subsection{Loop Closure Detection}
SLAM calculates the camera's position for the next frame based on the position of the previous frame, resulting in the accumulation of errors. Loop closure detection is introduced to eliminate the drift. Our system is divided into four stages: loop closure candidate frame detection, $sim3$ change calculation, loop closure fusion and pose optimization. For the current key frame $K_{i}$, we find the best three loop closure candidates and fusion candidates $K_{m}$ based on the similarity. Then for each candidate frame, we define a localized window containing 5 keyframes with the highest covariance with the candidate frame and the map points. To avoid false positives, the loop closure candidate frames are further subjected to covariance geometry checking as well as timing geometry checking. After that, the loop closure candidate frames selected after verification are used to construct a local window with the current keyframes for map fusion. Finally, the keyframes are optimized to eliminate the drift during tracking.
\subsection{Ablation Study}
In this section, we evaluate the choice of our depth estimation module, multivariate Gaussian encoding, depth loss term, and reparameterization.
\\
\\
\textbf{Depth Estimation}\quad We investigate the effect of the depth prediction module and depth distillation on the performance of our system. Table \ref{table4:Depth} shows the reconstruction performance of the system in monocular mode with the depth estimation module removed and the depth estimation module incorporated. Without the depth estimation module, our system suffers a dramatic drop in both tracking and mapping performance. However, with the cooperation of the depth estimation module and depth distillation module, the monocular mode performance of our system closely approaches that of the RGB-D mode. \\
\\
\textbf{Losses}\quad We further evaluate the impact of each loss on the reconstruction performance of the system. Table \ref{table3:headings} shows the impact of each loss on the reconstruction performance of the system, and it's easy to find that only by relying on $\mathcal{L}_{c}$ and $\mathcal{L}_{depth}$ cannot lead to the best reconstruction result. $\mathcal{L}_{sdf}$ has a more significant impact on the improvement of reconstruction quality because it affects the expression and reconstruction of the object's geometric surface information. In addition, $\mathcal{L}_{eik}$ and $\mathcal{L}_{dist}$ also have a positive impact on improving the reconstruction performance, one of them is used to optimize the geometric surface and the other is used to eliminate artifacts. Integrating all terms produces the best result.\\
\\
\textbf{Unbounded Scene}\quad We also conduct tests to assess the impact of reparameterization Equ. \ref{Equation:Unbound} and Gaussian encoding Equ. \ref{sigama} on reconstructing scenes with no defined boundaries. For comparison, we used NEWTON \cite{Matsuki2023NEWTONNV} as a baseline to evaluate our system's performance. The quantitative data in Table \ref{table5:unbounded} indicates that our reparameterization and Gaussian encoding significantly improve the accuracy in both scaling and geometric reconstruction of unbounded scenes. These approaches notably enhance our system's overall performance. Our data were acquired using the RGB-D mode of our system within the ScanNet 0207 scene. NEWTON's data are from the original paper \cite{Matsuki2023NEWTONNV}. The results on the ScanNet \cite{Dai2017ScanNetR3} indicate that our model shows greater robustness and more precise mapping abilities compared to NEWTON.\\
\\
\textbf{Camera Tracking}\quad We explore the effect of the depth loss term on camera localization. To verify the robustness of our proposed depth loss term, we further explore the results of solely applying L1 loss or L2 loss on the system. Table \ref{table7:headings} shows the experimental results on the Replica dataset \cite{Straub2019TheRD}. Our proposed depth loss term can significantly improve the camera tracking accuracy of the system. The convergence performance of L1 loss is poor, making it unsuitable for real-time SLAM systems. L2 loss is overly sensitive to depth outliers, hence not suitable for our depth estimation module-based SLAM system.
\begin{table}[t]
\begin{center}
\caption{\textbf{The impact of various losses}. All data are averaged over three runs in monocular mode on our system under seven scenes of the Replica dataset. '1' represents use this loss while '0' means ignore.}
\scriptsize
\renewcommand{\arraystretch}{1.5}
\setlength{\tabcolsep}{4.8mm}{
\begin{tabular}{l l l l l l l}
\hline
$\mathcal{L}_{c}$ &  $\mathcal{L}_{dep}$ & $\mathcal{L}_{eik}$ & $\mathcal{L}_{sdf}$ & $\mathcal{L}_{dist}$ & \makecell{Avg. F-score$\uparrow$} & PSNR$\uparrow$\\
\hline
\makecell{1} & \makecell{0} & \makecell{0} & \makecell{0} & \makecell{0} & \makecell{34.33}& \makecell{15.78}\\
\cdashline{1-7}[0.3pt/5pt]
\makecell{1} & \makecell{1} & \makecell{0} & \makecell{0} & \makecell{0} & \makecell{85.79} & \makecell{27.76}\\
\cdashline{1-7}[0.3pt/5pt]
\makecell{1} & \makecell{1} & \makecell{1} & \makecell{0} & \makecell{0} & \makecell{85.98} & \makecell{27.89}\\
\cdashline{1-7}[0.3pt/5pt]
\makecell{1} &\makecell{1} & \makecell{1} & \makecell{1} & \makecell{0} & \makecell{89.01} & \makecell{28.13}\\
\cdashline{1-7}[0.3pt/5pt]
\makecell{1} & \makecell{1} & \makecell{1} & \makecell{1} & \makecell{1} & \makecell{\pmb{89.73}} & \makecell{\pmb{28.37}}\\
\hline
\noalign{\smallskip}
\end{tabular}}
\label{table3:headings}
\end{center}
\end{table}
\begin{table}[!t]
\noindent
\begin{minipage}{\textwidth}
\scriptsize
\renewcommand{\arraystretch}{1.5}
\setlength{\tabcolsep}{2.5mm}
    \begin{minipage}[t]{0.45\textwidth}
        \captionsetup{skip=3pt}
        \caption{\textbf{Depth modules ablation study}. We explore the effects of the depth estimation module (DepE.) and the depth distillation module (DepD.) in our system. }
        \begin{tabular}{lcc}
            \hline
            & ATE[cm]$\downarrow$  & F-score$\uparrow$ \\ \hline
            None  & 0.46  & 83.79 \\
            DepE.  & 0.39  & 87.96 \\
            DepE./DepD.  & 0.35  & 89.43 \\
            RGB-D  & 0.33  & 90.96\\
            \hline
        \end{tabular}
        \raggedright
        \label{table4:Depth}
    \end{minipage}\hfill
    \begin{minipage}[t]{0.42\textwidth}
        \captionsetup{skip=3pt}
        \caption{\textbf{Unbounded scene ablation study}. We investigate the effects of the constructed function (Con.) and Gaussian encoding (Gau.) in our system.}
        \begin{tabular}{lcc}
            \hline
            & PSNR$\uparrow$ & SSIM$\uparrow$ \\ \hline
            NDC  & 19.09  & 0.607 \\
            NEWTON\cite{Matsuki2023NEWTONNV}  & 23.15  & 0.702 \\
            Con.  & 23.85  & 0.749 \\
            Con-Gau.  & 25.02  & 0.787\\
            \hline
        \end{tabular}
        \raggedright
        \label{table5:unbounded}
    \end{minipage}
\end{minipage}
\end{table}

\begin{table}[!t]
\caption{\textbf{Camera Tracking Ablation Study}. We report ATE RMSE[cm] on the Replica \cite{Straub2019TheRD}. Experimental results demonstrate that the loss function proposed by us exhibits superior robustness.}
\vspace{-3mm}
\begin{center}
\scriptsize
\renewcommand{\arraystretch}{1.5}
\setlength{\tabcolsep}{2.2mm}
\begin{tabular}{l l l l l l l l l l}
\hline\noalign{\smallskip}
 & \makecell{office0} & \makecell{office1} & \makecell{office2} &  \makecell{office3} & \makecell{office4}& \makecell{room0}& \makecell{room1}& \makecell{room2}& \makecell{\pmb{Avg.}}\\
\hline
\noalign{\smallskip}
 None& \makecell{0.310} & \makecell{0.327} & \makecell{0.339} & \makecell{0.512} & \makecell{0.603} & \makecell{0.588} & \makecell{0.391}& \makecell{0.307}& \makecell{0.422}\\
 \cdashline{1-10}[0.3pt/5pt]
 L1 Loss& \makecell{0.303} & \makecell{0.307}& \makecell{0.325}& \makecell{0.421}& \makecell{0.479}& \makecell{0.533} & \makecell{0.382}& \makecell{0.292}& \makecell{0.380}\\
  \cdashline{1-10}[0.3pt/5pt]
 L2 Loss& \makecell{0.307} & \makecell{0.295}& \makecell{\pmb{0.293}}& \makecell{0.414}& \makecell{0.527}& \makecell{\pmb{0.489}} & \makecell{0.407}& \makecell{0.291}& \makecell{0.377}\\
 \cdashline{1-10}[0.3pt/5pt]
 \textbf{Ours}& \makecell{\pmb{0.282}} & \makecell{\pmb{0.287}} & \makecell{0.302}& \makecell{\pmb{0.398}}& \makecell{\pmb{0.447}} & \makecell{0.502}& \makecell{\pmb{0.314}}& \makecell{\pmb{0.273}}& \makecell{\pmb{0.351}}\\
\hline
\noalign{\smallskip}
\end{tabular}
\label{table7:headings}
\end{center}
\end{table}

\begin{table}[!t]
\caption{\textbf{ATE RMSE [cm] on the TUM RGB-D dataset \cite{Sturm2012ABF}}. The performance metrics reported by our system for each scene are derived from the average of three executions. In contrast to other SLAM methods, our method outperforms in tracking accuracy.}
\vspace{-3mm}
\begin{center}
\scriptsize
\renewcommand{\arraystretch}{1.5}
\setlength{\tabcolsep}{6.0mm}
\begin{tabular}{l l l l l l l l}
\hline\noalign{\smallskip}
 &  Scene & \multicolumn{2}{c}{fr1-desk} & \multicolumn{2}{c}{fr2-xyz} & \multicolumn{2}{c}{fr3-office}\\
\hline
\noalign{\smallskip}
 & GO-SLAM\cite{Zhang2023GOSLAMGO} & \multicolumn{2}{c}{1.500} & \multicolumn{2}{c}{\pmb{0.600}} & \multicolumn{2}{c}{1.000}\\
 \cdashline{2-8}[0.3pt/5pt]
  RGB-D & Co-SLAM\cite{Wang2023CoSLAMJC} & \multicolumn{2}{c}{2.400} & \multicolumn{2}{c}{1.700} & \multicolumn{2}{c}{2.400}\\
  \cdashline{2-8}[0.3pt/5pt]
 & \textbf{Ours} & \multicolumn{2}{c}{\pmb{1.382}} & \multicolumn{2}{c}{0.641} & \multicolumn{2}{c}{\pmb{0.939}}\\
 \hline\noalign{\smallskip}
 & DROID-SLAM\cite{Teed2021DROIDSLAMDV} & \multicolumn{2}{c}{2.713} & \multicolumn{2}{c}{1.304} & \multicolumn{2}{c}{2.573}\\
 \cdashline{2-8}[0.3pt/5pt]
 & DeepV2D\cite{Teed2018DeepV2DVT}& \multicolumn{2}{c}{21.391} & \multicolumn{2}{c}{6.950} & \multicolumn{2}{c}{11.922}\\
 \cdashline{2-8}[0.3pt/5pt]
Mono & ORB-SLAM3\cite{Campos2020ORBSLAM3AA} & \multicolumn{2}{c}{1.534} & \multicolumn{2}{c}{0.720} & \multicolumn{2}{c}{1.400}\\
\cdashline{2-8}[0.3pt/5pt]
 & GO-SLAM\cite{Zhang2023GOSLAMGO} & \multicolumn{2}{c}{2.381} & \multicolumn{2}{c}{1.093} & \multicolumn{2}{c}{1.773}\\
 \cdashline{2-8}[0.3pt/5pt]
 & \textbf{Ours} & \multicolumn{2}{c}{\pmb{1.481}} & \multicolumn{2}{c}{\pmb{0.686}} & \multicolumn{2}{c}{\pmb{1.101}}\\
\hline
\noalign{\smallskip}
\end{tabular}
\label{table8:headings}
\end{center}
\end{table}
\subsection{Additional Experimental Results}
\begin{figure}[!ht]
\centering
\includegraphics[width=120mm]{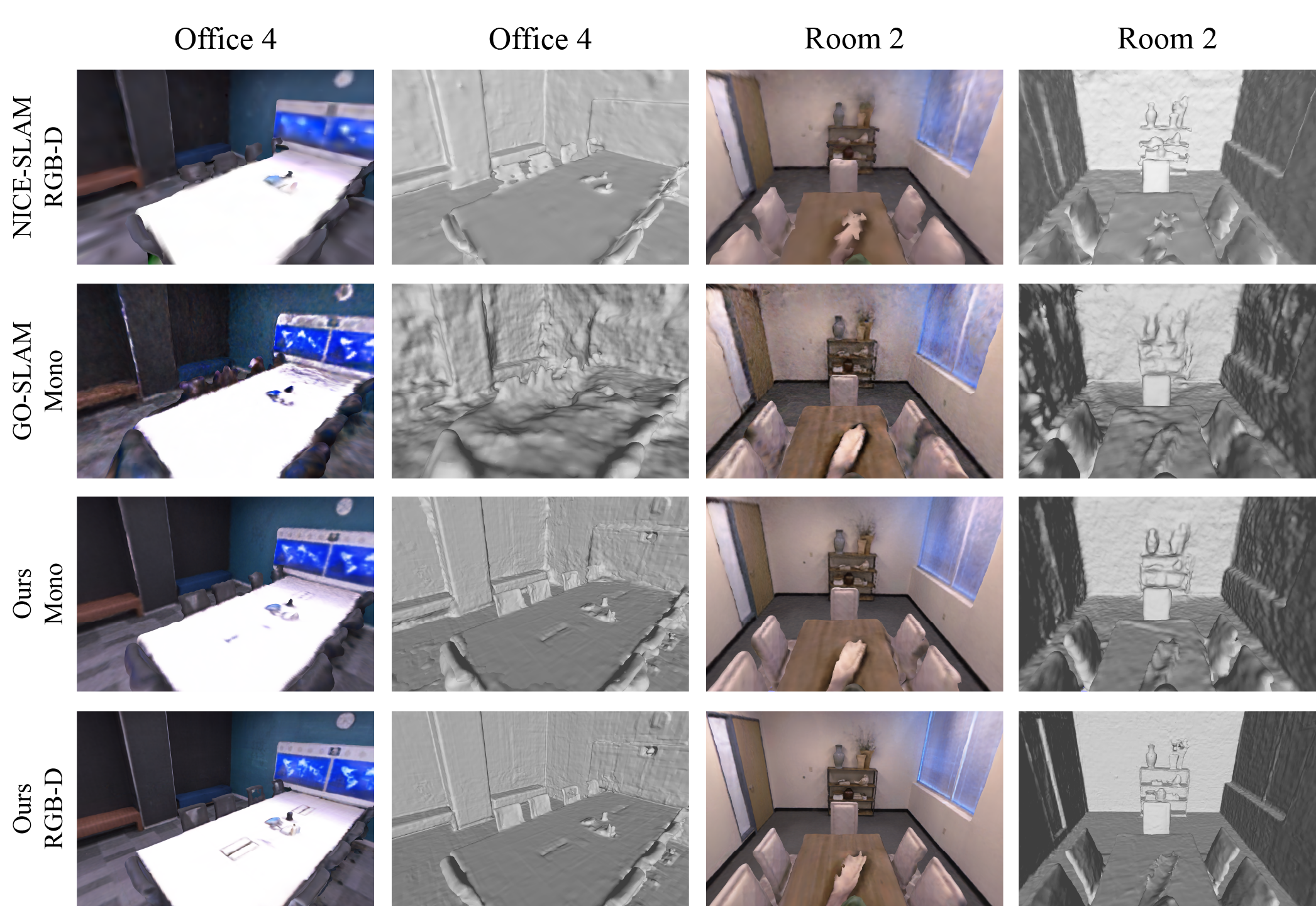}
\captionsetup{justification=justified,singlelinecheck=false}
\caption{\textbf{Reconstruction results on Replica \cite{Straub2019TheRD}}. Qualitative comparison on Replica office4 and room2. }
\label{fig8:Replica}
\end{figure}

\begin{figure}[!t]
\centering
\includegraphics[width=120mm]{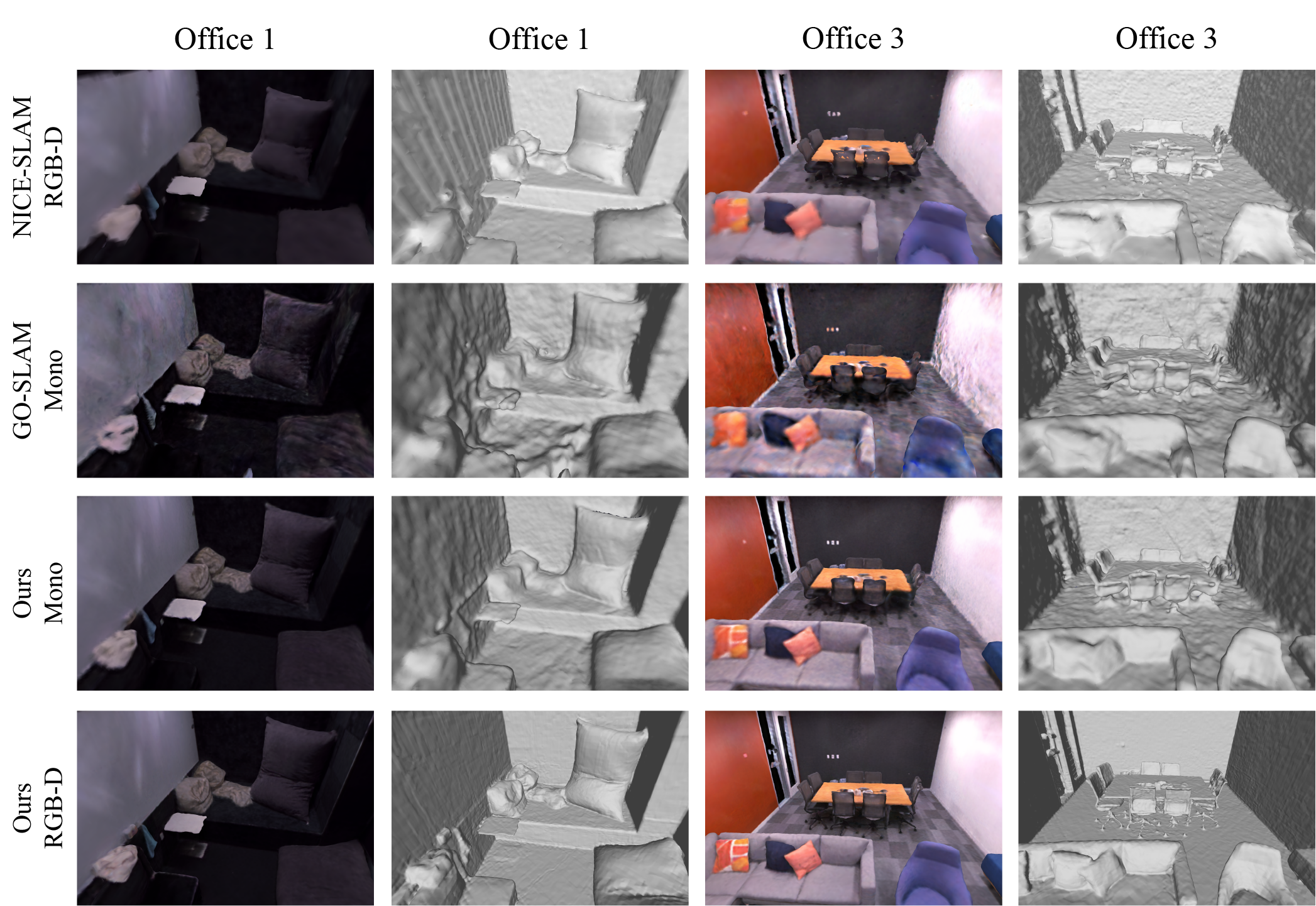}
\captionsetup{justification=justified,singlelinecheck=false}
\caption{\textbf{Reconstruction results on Replica \cite{Straub2019TheRD}}. Qualitative comparison on Replica office1 and office3.  }
\label{fig9:Replica}
\end{figure}

\textbf{Evaluation on TUM-RGBD \cite{Sturm2012ABF}}\quad To test our method further, we compared our method with point-cloud SLAM methods \cite{Teed2021DROIDSLAMDV, Teed2018DeepV2DVT, Campos2020ORBSLAM3AA} and NeRF-based SLAM methods \cite{Zhang2023GOSLAMGO, Wang2023CoSLAMJC} on the TUM RGB-D dataset. Since the TUM RGB-D dataset lacks ground truth meshes, we compared the accuracy of the tracking process. Table \ref{table8:headings} shows the accurate tracking performance of our method.\\
\\
\textbf{Evaluation on Replica \cite{Straub2019TheRD}}\quad Here we show more results on Replica \cite{Straub2019TheRD}. Fig. \ref{fig8:Replica}, Fig. \ref{fig9:Replica} and demonstrate the superior scene reconstruction capability of our proposed method in monocular and RGB-D mode, and also exhibiting enhanced smoothness in comparison to GO-SLAM \cite{Zhang2023GOSLAMGO}.

\end{document}